\documentclass[10pt,twocolumn,letterpaper]{article}

\usepackage{cvpr}              

\usepackage{graphicx}
\usepackage{amsmath}
\usepackage{amssymb}
\usepackage{booktabs}

\usepackage[pagebackref,breaklinks,colorlinks]{hyperref}

\usepackage[capitalize]{cleveref}
\crefname{section}{Sec.}{Secs.}
\Crefname{section}{Section}{Sections}
\Crefname{table}{Table}{Tables}
\crefname{table}{Tab.}{Tabs.}

\def\cvprPaperID{*****} 
\def\confName{CVPR}
\def\confYear{2023}
\begin{document}

\title{Technical Report for Argoverse Challenges on 4D Occupancy Forecasting}
\author{Pengfei Zheng \thanks{Equal contribution} \thanks{Work done as an intern at Lenovo Research.}\label{intern} \textsuperscript{1}, Kanokphan Lertniphonphan\footnotemark[1] \textsuperscript{3}, Feng Chen \footnotemark[1] \textsuperscript{3}, Siwei Chen \footnotemark[\value{footnote}] \textsuperscript{2},\\ Bingchuan Sun\textsuperscript{3}, Jun Xie\textsuperscript{3}, Zhepeng Wang\textsuperscript{3}\\
\textsuperscript{1}University of Science and Technology Beijing,
\textsuperscript{2}Tsinghua University\\
\textsuperscript{3}Lenovo Research\\
{\tt\small {chenfeng13}@lenovo.com}
}

\maketitle
\begin{abstract}
This report presents our Le3DE2E\_Occ solution for 4D Occupancy Forecasting in Argoverse Challenges at CVPR 2023 Workshop on Autonomous Driving (WAD). Our solution consists of a strong LiDAR-based Bird's Eye View (BEV) encoder with temporal fusion and a two-stage decoder, which combines a DETR head and a UNet decoder. The solution was tested on the Argoverse 2 sensor dataset to evaluate the occupancy state 3 seconds in the future. Our solution achieved 18\% lower L1 Error (3.57) than the baseline on the 4D Occupancy Forecasting task in Argoverse Challenges at CVPR 2023.
\end{abstract}
\section{Introduction}
\label{sec:intro}

4D occupancy forecasting is a new challenge introduced in CVPR 2023 WAD. The task is to understand how an environment evolves with time which is crucial for motion planning in autonomous driving. However, conventional methods require costly human annotations, such as detection bounding boxes, tracking ids, or semantic segmentation, which make it difficult to scale up to a large labeled dataset. This challenge aims to learn future occupancy forecasting from unlabeled datasets.   

In this challenge, given a particular agent's observation of the world in the past n seconds, we need to predict the space-time evolution of the world in the future n seconds. Specifically, the occupancy state of 5 frames in the next 3s is predicted by observing the point cloud of 5 frames in the past and present timestamp within 3s (at a frequency of 5/3Hz). The future frame point cloud is obtained by rendering the occupancy state from a given query ray. Then, all point clouds and occupancies are aligned to the current frame under the LIDAR coordinate system.

Our solution adopts a voxel feature encoder to transform LiDAR point clouds into a 2D Bird's Eye View (BEV) feature map and uses a DETR\cite{10.1007/978-3-030-58452-8_13} structure head to predict voxel-wise future occupancy. To refine the results, a UNet\cite{Ronneberger2015UNetCN} head is employed as a second-stage decoder to produce more accurate forecasting results.

\begin{figure*}[ht]
  \centering
   \includegraphics[width=\linewidth, height=3.5cm]{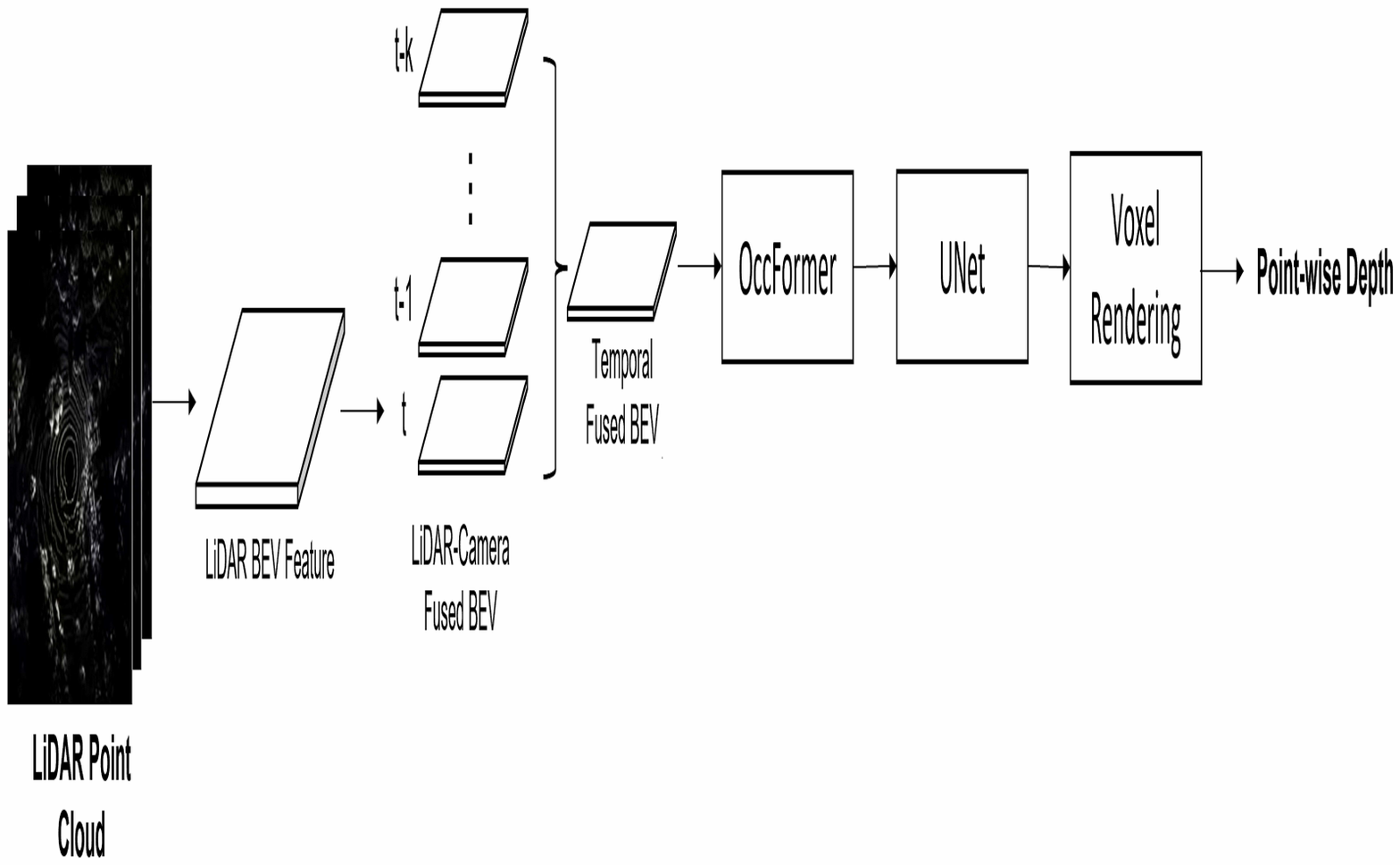}
   \caption{\textbf{System overview.} Firstly the LiDAR point clouds of the current frame are voxelized and encoded to the \textbf{BEV feature map}. Secondly, the historical frame BEV feature maps are fused with the current frame by a \textbf{temporal encoder}. Thirdly, Spatial-temporal fused BEV is fed into \textbf{OccFormer} which generates voxel-wise occupancy forecasting results. Fourthly, \textbf{UNet} works as a second-stage decoder to refine the forecasts. Last, \textbf{voxel rendering} generates point-wise depth along the given ray direction as post-processing. }
   \label{fig:system}
\end{figure*}

\begin{table*}
  \centering
  \begin{tabular}{c c c c c}
    \toprule
    Team & L1 ($\downarrow$) & AbsRel & NFCD & CD \\
    \midrule
    Host\_34336\_Team (Raytracing) & 6.68 & 0.48 & 8.79 & 49.88 \\
    Host\_34336\_Team (Point Cloud Forecasting as a P) & 4.34	& 0.26 & 5.23 & 92.08 \\
    \hline
    Le3DE2E\_Occ (Ours) & 3.57 & 0.22 & 3.35 & 91.61 \\
    \bottomrule
  \end{tabular}
  \caption{4D Occupancy Forecasting on the official leaderboard}
  \label{tab:result}
\end{table*}

\section{Method}
\label{sec:method}

We employ the bev feature as  a unified representation and adopted the OccFormer module as an occupancy head for 4d occupancy forecasting based on UniAD\cite{hu2023_uniad}. OccFormer module consists of a transformer decoder with T sequential blocks. The pipeline of our method is shown in Figure 1.

\subsection{BEV feature}
We employ the LIDAR BEV encoder based on SECOND\cite{yan2018second} to generate the LIDAR BEV feature \(B_l\). The LIDAR encoder takes as input the current and past \(T=5\) frames of the point cloud with a time step of 0.6 s. All point clouds have been aligned to the current frame. Afterward, we fuse the past T frames with the current frame to aggregate temporal features by a 2D convolutional block. The spatial-temporal fused BEV feature map is fed to the occupancy decoder.

\subsection{OccFormer}

Given the BEV feature from upstream modules, we feed it to OccFormer to get the multi-time-step occupancy results. OccFormer consists of \(T\) sequential blocks, where \( T=5\) is a number of future frames. Each block is responsible for generating the occupancy of a particular frame. Unlike the instance occupancy output by OccFormer in UniAD, we need dense voxel-wise occupancy. 

In each sequential block, the BEV features first perform self-queries through the self-attentive layer to compute the similarity metric within the feature. Then, we  randomly initialize the instance query \(Q_I\), which are track query \(Q_A\) , agent position \(P_A\) and motion query \(Q_X\). The instance query \(Q_I\) and BEV feature interact using cross-attention so that the dense scene feature and sparse agent feature benefit from each other. The interacted BEV feature is sent to the next sequential block, which is cycled in RNN fashion. 

All \(T\) frame BEV features are upsampled to obtain future occupancy \(O_T\). Dimensions of all dense features and instance features are 256 in OccFormer. 
A UNet head is used as a second-stage decoder to enhance the multiscale forecasting results. After generating voxel-wise occupancy forecasting, voxel rendering is then performed by the query rays to get the point-wise estimated depth.

\section{Experiments}
\label{sec:experiments}

\subsection{Dataset}

The competition used the Argoverse 2 Sensor Dataset\cite{Argoverse2}, which consisted of 1000 scenes (750 for training, 150 for validation, and 150 for testing) with a total of 4.2 hours of driving data. The total dataset is extracted in the form of 1 TB of data. Each vehicle log has a duration of approximately 15 seconds and includes an average of approximately 150 LiDAR scans with 10 FPS LiDAR frames.

\subsection{Evaluation Metrics}

The metric is performed in the range of [-70m, -70m, -4.5m, 70m, 70m, 4.5m] with a voxel resolution of 0.2m.  The absolute L1 distance between the true expected depth along a given ray direction and the predicted expected depth obtained by rendering along the same ray direction is used as the main metric. Absolute relative error (AbsRel), near-field chamfer distance (NFCD), and vanilla chamfer distance (CD) are measured together as other metrics. More details of the evaluation can be found in the paper \cite{khurana2023point}.

\subsection{Implementation details}

We followed the baseline \cite{khurana2023point} for data preparation for training and evaluation. Point shuffle is used for data augmentation in the training stage. The voxel size is set to (0.075m, 0.075m, 0.2m) and the resulting BEV feature \(B_l\) has a shape of 240*240. We train the model with a cosine annealing policy with a 0.001 learning rate and use L1 loss and Adam optimizer. We train the model from scratch for 20 epochs on 8 V100 GPUs with a total batch size of 8. 

We submitted our results to the testing server and got a 3.57 L1 Error. The results are shown in \autoref{tab:result}. Our results outperform the baseline with $18\%$ and $15\%$ improvements in L1 Error and Absolute Relative L1 Error, respectively.

\section{Conclusion}
\label{sec:conclusion}

In our model, we employ a LIDAR encoder to encode spatial features and then do a temporal fusion of historical BEV features. We use the BEV feature as a unified intermediate representation. We employ an OccFormer as 4D occupancy prediction head to loop out future occupancy in RNN style. Following the work of \cite{khurana2023point}, we "render" point cloud data from 4D occupancy predictions and estimate depth for supervision. The experimental results indicate that our model achieved better scores than the baseline with $18\%$ and $15\%$ improvements in L1 Error and Absolute Relative L1 Error, respectively.


{\small
\bibliographystyle{ieee_fullname}
\bibliography{egbib}
}

\end{document}